\documentclass[runningheads]{llncs}
\usepackage[T1]{fontenc}
\usepackage{graphicx}
\usepackage[hidelinks]{hyperref}
\usepackage{enumitem}
\usepackage{comment}
\usepackage{multirow}

\usepackage{color}

\usepackage{everypage}

\AddThispageHook{%
  \put(-10,20){%
    \parbox[t]{1.4\textwidth}{%
      \raggedright
      \footnotesize
      \textcolor{blue}{
This preprint has not undergone peer review (when applicable) or any post-submission improvements or corrections. \\
The Version of Record of this contribution is published in Lecture Notes in Computer Science, vol 16121,
and is available online at \texttt{https://doi.org/10.1007/978-3-032-05176-9\_{24}}.
}
    }%
  }%
}

\begin{document}
\title{SynDocDis: A Metadata-Driven Framework for Generating Synthetic Physician Discussions Using Large Language Models }
\titlerunning{Metadata-Driven Generation of Synthetic Physician Discussions}
%


\author{Beny Rubinstein\inst{1}\orcidID{0000-0001-6578-1467} \and
Sérgio Matos\inst{1,2}\orcidID{0000-0003-1941-3983}}
\authorrunning{B. Rubinstein and S. Matos}
%
\institute{University of Aveiro, Portugal\\
\and
IEETA, DETI, LASI, University of Aveiro, Portugal\\
\email{\{BenyR,aleixomatos\}@ua.pt}}

\maketitle              
\begin{abstract}
Physician-physician discussions of patient cases represent a rich source of clinical knowledge and reasoning that could feed AI agents to enrich and even participate in subsequent interactions. However, privacy regulations and ethical considerations severely restrict access to such data. While synthetic data generation using Large Language Models offers a promising alternative, existing approaches primarily focus on patient-physician interactions or structured medical records, leaving a significant gap in physician-to-physician communication synthesis. 

We present SynDocDis, a novel framework that combines structured prompting techniques with privacy-preserving de-identified case metadata to generate clinically accurate physician-to-physician dialogues. Evaluation by five practicing physicians in nine oncology and hepatology scenarios demonstrated exceptional communication effectiveness (mean 4.4/5) and strong medical content quality (mean 4.1/5), with substantial inter-rater reliability ($\kappa$ = 0.70, 95\% CI: 0.67-0.73). 
The framework achieved 91\% clinical relevance ratings, while maintaining doctors' and patients' privacy. These results place SynDocDis as a promising framework for advancing medical AI research ethically and responsibly through privacy-compliant synthetic physician dialogue generation with direct applications in medical education and clinical decision support. 

\end{abstract}
\section{Introduction}

The integration of large language models (LLMs) into medicine represents a paradigm shift, moving beyond task automation to a synergistic partnership between artificial intelligence and human expertise. LLMs have surpassed previous NLP systems in medical question answering, summarization, text generation, and clinical decision support~\cite{Peng2023}; yet, an area where research is lagging behind is supporting physician-to-physician discussions of patient cases. Progress in this area requires access to real patient case discussions, which is challenging given the highly sensitive nature of patient data and strict privacy regulations, such as HIPAA in the United States and GDPR in the EU. Furthermore, physician discussions remain sensitive even when de-identified, because of potential re-identification risks, and doctors are reluctant to share their decision-making processes due to fear of scrutiny and liability risks.  

This study contributes to the assessment of synthetic clinical text generation for physician-physician discussions as a balanced approach that allows data sharing with researchers while minimizing privacy risks. 
We present a framework called SynDocDis, which focuses on generating synthetic physician-to-physician discussions using LLMs. Its primary goal is to create clinically accurate and privacy-preserving synthetic dialogues based on de-identified metadata extracted from real patient case discussions while maintaining both clinical accuracy and the natural communication patterns of discussions among physicians.
This work is motivated by challenges in accessing real-world data for physician conversations due to privacy concerns, liability issues, and the sensitivity of de-identified discussions, and addresses two key research questions:

\begin{enumerate}
    \item \textit{Can general-purpose large language models (LLMs) generate high-fidelity physician-to-physician discussions using only de-identified metadata from real patient case discussions?}
    \item \textit{How do domain experts judge the quality of these synthetic discussions?}

\end{enumerate}

The contributions of this study can be summarized as follows:

\begin{itemize}
    \item We propose SynDocDis, a generic framework based on Context-Instructions-Details-Input (CIDI)\footnote{The CIDI framework is articulated in the LinkedIn course \href{https://www.linkedin.com/learning/nano-tips-for-using-chat-gpt-to-10x-your-productivity-at-work-with-gianluca-mauro/from-casual-to-pro-how-advanced-chatgpt-usage-looks-like}{``Nano Tips for Using Chat GPT to 10x Your Productivity at Work with Gianluca Mauro''}}  for generating physician discussions using metadata;
    \item We conduct a comprehensive evaluation of nine synthetic case discussions by practicing physicians from diverse specialties and geographical locations;
    \item We share a public package containing the prompts used, the synthetic medical dialogues, and expert ratings: \url{https://anonymous.4open.science/r/syndocdis}

\end{itemize}

\section{Related Work}

Recent advances in clinician text-data generation have demonstrated significant potential for addressing data scarcity and privacy concerns in medical AI research. The current literature explores various facets of how LLMs can generate high-fidelity medical text, but there is limited direct evidence that they can reliably create authentic physician-to-physician discussions using only de-identified metadata from real patient-case discussions.

\subsection{LLM Evolution in Clinical Text Generation}

LLMs have rapidly advanced the generation of synthetic medical dialogue, particularly in tasks such as clinical text summarization (e.g., discharge summaries), medical question answering, and patient-provider communication. For example, Sufi~\cite{Sufi2024} extracted features from synthetically generated patient discharge messages that correlated with the severity of  patients' conditions. Moreover, research by Williams et al.~\cite{Williams2025} revealed that LLM-generated discharge summaries can meet quality and safety standards comparable to those written by human physicians, albeit with slightly higher error rates and lower comprehensiveness. 
Other studies have shown that LLM-generated text can outperform human responses in terms of quality, capturing nuanced communications that reflect both technical proficiency and empathy. Ayers et al.~\cite{Ayers2023} highlighted that AI chatbots can generate responses that show quality and empathy for medical advice. This aligns with another study by Hartman et al.~\cite{Hartman2024}, which suggested that LLMs could streamline clinical tasks, such as emergency handoff notes, thus alleviating the burdens faced by health professionals and fostering effective communication between physicians. Their findings indicate the superiority of LLM-generated over physician-written notes in accuracy and detail, but show that they are marginally inferior in terms of usefulness and safety, suggesting that a physician-in-the-loop implementation design is necessary for safe adoption. Moreover, the clinical potential of LLMs in this context has been growing with their increasing capability to provide diagnostic support and reasoning explanations, which are crucial components of physician-to-physician dialogue. Spitzer et al.~\cite{Spitzer2025} note that LLMs can assist in generating explanatory content that clarifies the rationale behind clinical decisions, thereby enriching peer discussions and collaborative patient care. Such capabilities are central to creating more fluid communication between physicians.
Although the implementation of LLMs in clinical settings is promising, several challenges remain. Small et al.~\cite{Small2024} addressed concerns related to processing needs, model biases, and privacy, all of which hinder the deployment of LLM technology in  clinical settings. Nonetheless, advancements in these models suggest that with continued refinement and resolution of these issues, LLM-based agents may complement and facilitate detailed and accurate case discussions.

\subsection{Synthetic Medical Dialogue}

The synthesis of medical dialogues represents a particularly challenging subset of medical data generation, which requires the preservation of both clinical accuracy and natural conversation flow. Existing research has primarily focused on patient-physician interactions; for example, a recent study indicated that although patients often express a preference for human doctors over AI, there is interest in hybrid models that incorporate AI assistance during consultations~\cite{Riedl2024}. 
Recent literature also suggests that generative AI can transform the clinician-patient relationship into a triadic interaction that includes AI as a ``third agent,'' ultimately guiding shared decision-making processes~\cite{CamposJr2024,Lorenzini2023}.  

Previous studies have explored various aspects of medical dialogue generation, including record-centric narrative synthesis and privacy-preserving techniques for sharing healthcare data. Moser et al.~\cite{Moser2024} highlight the potential of LLMs to generate realistic medical dialogues, which is crucial for replicating natural conversational nuances in the clinical setting. This was complemented by Li et al.~\cite{Li2021}, who focused on semi-supervised variational reasoning for dialogue generation, which embodies strategies that could enhance the quality of inter-physician interactions by simulating inquiry and dialogue management, typical of clinical practice.
In the context of dataset generation, Liu et al.~\cite{Liu2022} emphasized the importance of various medical conversational datasets aimed at enhancing dialogue agents that assist healthcare professionals. Similarly, Zeng et al.~\cite{Zeng2020} introduced MedDialog, a dataset containing over three million turns of patient-doctor conversations in English and Chinese collected from online medical consultation platforms. This demonstrated that large-scale, privacy-preserved dialogue datasets can significantly improve the performance of medical dialogue systems, particularly in maintaining clinical accuracy and natural conversation flow.

While LLMs have made strides in synthesizing clinical text, such as discharge summaries or single-speaker clinical notes, there is a gap in research on their ability to create high-fidelity, authentic physician-to-physician dialogues based solely on de-identified metadata from real patient cases. This gap is due to the lack of  conversational context, complexity of multispeaker interactions, and insufficient validation in this specific scenario. Multispeaker dialogue synthesis poses unique challenges related to conversational dynamics, speaker roles, and interaction contexts that are not present in standard clinical text summarization. 

\subsection{Gaps in Current Approaches}

Despite the advances highlighted in the previous sections, several significant gaps remain, including:

\begin{itemize}
    \item Limited focus on physician-to-physician communication: although substantial work exists on patient-provider dialogues, the synthesis of physician-to-physician discussions remains largely unexplored. This is particularly significant, given the unique characteristics of interprofessional medical communication, including specialized terminology and complex clinical reasoning;
    \item Privacy-utility trade-offs: existing approaches often sacrifice utility for privacy preservation, particularly when generating detailed clinical discussions. Giuffrè and Shung highlighted these challenges, but did not resolve them for dialogue generation~\cite{Giuffr2023};
    \item Evaluation metrics: while current evaluation approaches are adequate for structured data, they lack specific metrics for assessing the quality of synthetic medical dialogues, particularly in professional-to-professional communication.
\end{itemize}

Our approach builds upon existing work while addressing the unique challenges of synthetic physician-to-physician discussions, contributing to both the theoretical understanding and practical implementation of synthetic medical data generation. To our knowledge, our work is the first to (i) condition generation on shareable, de-identified metadata, (ii) create multi-speaker discussions with role tags and reference stubs, and (iii) evaluate them with practicing physicians from four specialties.
In summary, the evidence suggests that general-purpose LLMs can generate high-fidelity physician-to-physician discussions by leveraging de-identified metadata from real patient cases.

\section{Methods}

We developed a new method for generating synthetic physician discussions, focusing on preserving the main characteristics of real-world medical conversations. 
We hypothesized that by carefully designing prompts to guide the models, we could create realistic and relevant discussions that protect privacy and are useful for developing AI tools to support physician discussions.

\subsection{Study Design}

\begin{figure}
\centering
\includegraphics[width=\linewidth]{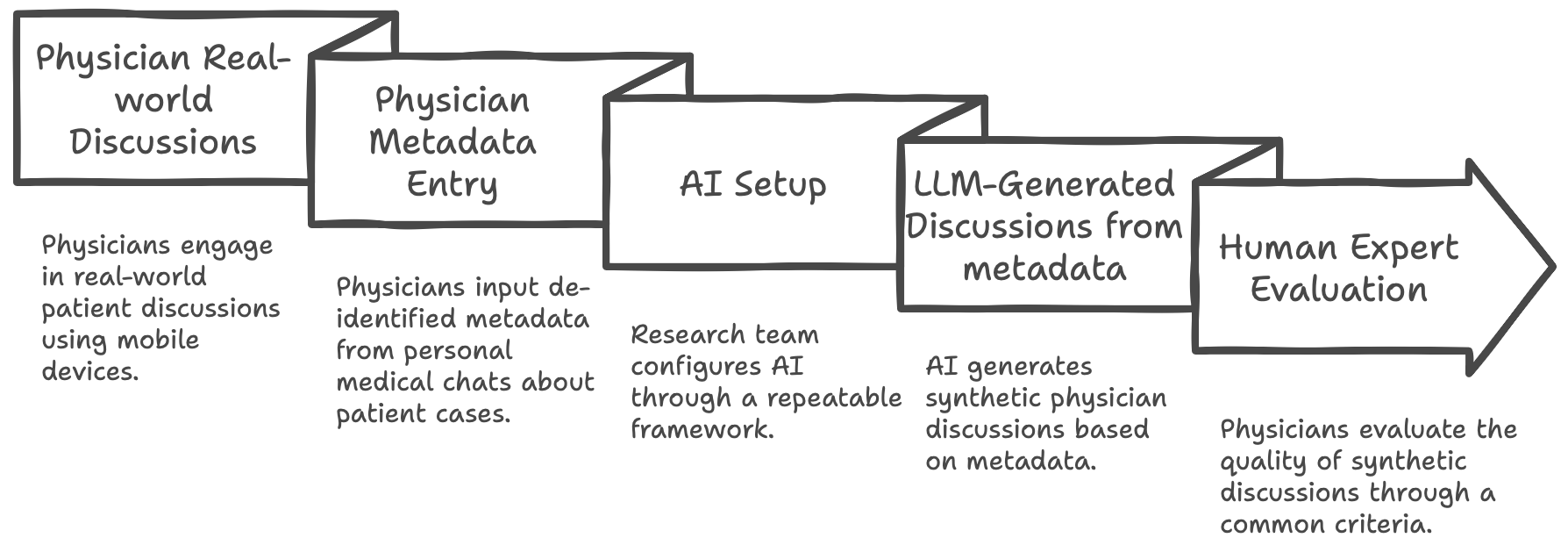}
\caption{Overview of study design and workflow.}
\label{fig:workflow}
\end{figure}

Our approach focuses on three key aspects: (a) input: de-identified metadata extraction from real-world physician discussions; (b) setup: implementation of the context-instructions-details-input (CIDI) framework for structured prompting; and (c) output evaluation: expert physician evaluation of synthetic data quality. The framework uses metadata from real-world physician discussions to generate synthetic expert discussions that maintain privacy while preserving clinical relevance and similar dynamics of medical discussions, which go through a structured human evaluation process (see Fig.~\ref{fig:workflow}). 
In addition, we aimed to generate diverse responses whenever possible because physicians value the variability of perspectives that are not always present in their real-world discussions, and also to include references to claims made in these simulated discussions about patient cases to ensure that they are evidence-based.

\subsection{Data Collection: Physician input through metadata}

We developed a data entry form and asked physicians to collect metadata describing real patient case discussions they were engaged in through their professional medical chat (PMC) groups (see Table~\ref{tab:metadata}).
We then used these metadata as input to create high-fidelity synthetic data.

\begin{table}
  \caption{Example of metadata provided from real-world physician discussions.}
  \label{tab:metadata}
  \begin{tabular}{|p{0.3\linewidth} | p{0.65\linewidth}|}
    \hline
    Metadata&Example\\
    \hline
    Chat Name & Pancreas ATM\\
    Participant Doctors & 52, 35, 6, 65, 46\\
    Specialty of Participants & Head of Department in a large hospital; Head of Service in a peripheral hospital; Head of Unit in a large hospital; Head of Unit in a large hospital; Head of Unit in a large hospital.\\
    Patient Case & Male, age 69, PS1, Aug 2023 diagnosed with pancreatic adenocarcinoma, one lesion in pancreatic head and another in tail. Underwent total pancreatectomy, pathologic stage T2mN0M0, now is on adjuvant FOLFIRINOX. Genetic testing shows germinal mutation ATM C.103C>T heterozygote. The question is whether to give him maintenance PARP inhibitor.\\
    Number of responses & 4\\
    Are answers valuable? & Valuable. Standard of care\\
    Variability in responses & All replied the same laconic- against PARP inhibitor. One added there is only evidence in BRCA mut.\\

    \hline
\end{tabular}
\end{table}

\subsection{Synthetic Data Generation Framework}

\begin{figure}
\centering
\includegraphics[width=\linewidth]{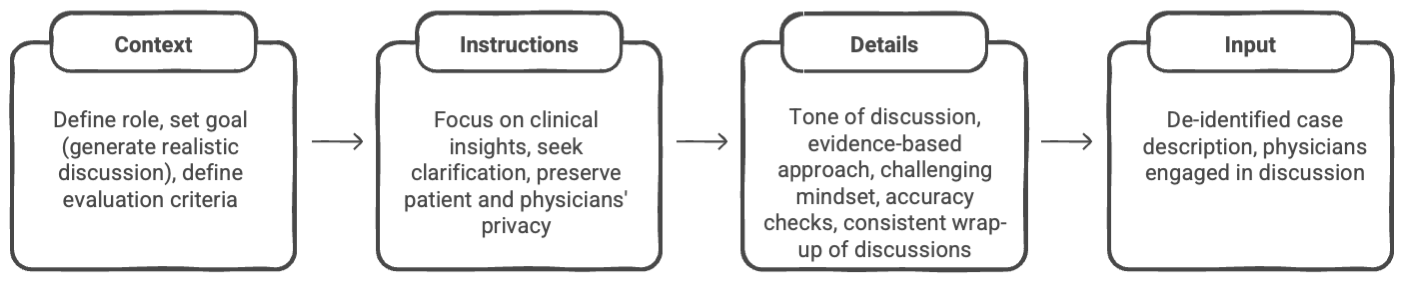}
\caption{Generation of synthetic physician discussions following the CIDI framework.}
\label{fig:cidi}
\end{figure}

We applied the ``Context-Instructions-Details-Input'' (CIDI) framework to structure the generation of synthetic physician discussions based on de-identified case metadata to establish a repeatable process and to enhance both the specificity and relevance of the AI-generated discussions (Fig.~\ref{fig:cidi}).

\subsubsection{User Prompting} In addition to role-playing---asking the model to adopt a persona and act accordingly---we adopted a technique called emotion prompting, using capital letters to emphasize important aspects (see Fig.~\ref{fig:system_prompt}, top).

\subsubsection{System Prompting} We applied prompting techniques such as role-playing to set up the context, and a hybrid prompting structure---mainly a zero-shot approach with a chain-of-thought technique---to instruct the LLM to approach the task step-by-step, followed by a definition of the expected output (Fig.~\ref{fig:system_prompt})~\cite{Savage2024,wei2022chain}.

\begin{figure}
\centering
\fbox{\begin{minipage}{0.95\linewidth}
\linespread{1.05}\selectfont
\textbf{User Prompt:} Act as an experienced and helpful physician and Head of a Unit/Department in a large hospital. With extensive experience and thousands of patient cases, you also moderate a WhatsApp group for oncologists with broad general medical knowledge. It is ESSENTIAL to create the discussion based on the provided INPUT which you will request from me, aiming for a thoughtful, high-value exchange among expert peers. Focus on enriching the conversation with clinical insights, expertise, and best practices in oncology and general medicine. Ensure that P0 (the `Case owner') actively addresses relevant questions or contributions from other doctors as part of the patient case discussion.
\end{minipage}}

\vspace{0.5cm}

\fbox{\begin{minipage}{0.95\linewidth}
\linespread{1.05}\selectfont
\textbf{System Prompt:} You are a knowledgeable and experienced physician specializing in oncology. Your goal is to generate realistic discussions between physicians about patient cases based on the metadata that I will provide. Please maintain a professional, clear, and concise tone, with each doctor focusing on relevant clinical insights, differential diagnoses, and actionable next steps. Doctors will seek clarification, agree with, or build upon others’ suggestions when appropriate.

\textbf{Instructions:} <begin>

• Step 1: Create a discussion that combines supportive, exploratory, and question-driven responses. Include occasional clarifications, agreements, challenges, or new and alternative suggestions to mirror real-life physician discussions. Ensure 'Case owner' participates by addressing questions, adding insights, asking clarifying questions, or responding to clarifications throughout the discussion as needed.

• Step 2: State the total number of unique physicians who participated in the discussion, counted once per doctor (and list their numbers). For example, if I enter 35, 10, 14 as input to <Physicians participating> then you should state ``3 physicians engaged in the discussion: P35, P10, and P14''

• Step 3: Confirm the total number of replies (R) in the discussion, including all contributions from the 'Case owner' and other physicians as per the specified response count. For example, if 'Case owner' posted a case, and each of the additional two physicians responded twice, then R=2 physicians x 2 responses per physician (so R=4).

• Step 4: Cite relevant external references, such as research papers or medical guidelines, to enhance the clinical depth of the discussion when applicable. <end>''

\textbf{Output:} Begin the discussion with 'Case owner' presenting the case details and initial questions to peers. Conclude the conversation once all responses and follow-up from 'Case owner' have been addressed. Name each responding physician in the format ``Doctor'' followed by the number of the physician.  Keep responses concise and relevant, each on a new line, ensuring they logically build on previous inputs. Adhere to the specified count for both unique physicians (P) and total responses (R).

\textbf{Reward Criteria:} An external physician will receive this guideline to assess output based on the following criteria:\\
1.Medical Content Quality Assessment\\
1.1 Clinical Accuracy: Assess whether the information shared is medically correct and up to date.\\
(…)\\
2.4 Variability of Responses: Evaluate if physicians are challenging each other and exploring a diverse set of ``schools of thoughts.''

\end{minipage}}
\caption{User and System prompts used to generate synthetic patient case discussions.}
\label{fig:system_prompt}
\end{figure}

\begin{table}
\caption{Evaluation criteria used to assess the synthetic discussions.}
\label{tab:criteria}

\newlength{\firstColActualWidth}
\setlength{\firstColActualWidth}{0.7cm} 

\newlength{\parboxWrapWidth}
\setlength{\parboxWrapWidth}{3cm} 

\begin{tabular}{|p{\firstColActualWidth}|p{0.3\linewidth} | p{0.6\linewidth}|}
\hline
\multicolumn{1}{|c|}{} & Category & Description \\
\hline
\multirow{4}{*}{\centering 
    \rotatebox[origin=c]{90}{
            \parbox[c][\firstColActualWidth][c]{\parboxWrapWidth}{%
            \centering 
            \strut Medical Content Quality\strut
        }%
    }%
} & 1.1 Clinical Accuracy & Assess whether the information shared is medically correct and up to date.\\
    &1.2 Evidence-Based & Evaluate if physicians are using evidence-based medicine principles in their discussion.\\
    &1.3 Relevance & Is the discussion directly applicable to the patient case at hand?\\
    &1.4 Comprehensiveness & Check if all relevant aspects of the patient's condition are being addressed.\\

\multirow{4}{*}{\centering
    \rotatebox[origin=c]{90}{%
        \parbox[c][\firstColActualWidth][c]{\parboxWrapWidth}{%
            \centering
            \strut Communication Effectiveness\strut%
        }%
    }%
} & 2.1 Clarity \& Coherence & Assess whether the discussion is clear, well-structured, and easy to follow.\\
    &2.2. Medical Terminology & Evaluate appropriate use of medical terms and explanations when necessary.\\
    &2.3 Active Listening & Observe if physicians are attentively listening to each other's input. \\
    &2.4 Variability/Diversity & Evaluate if physicians are challenging each other and exploring a diverse set of ``schools of thought''.\\
\hline
\end{tabular}
\end{table}

\subsection{Evaluation Protocol}

Our evaluation protocol for the quality of synthetic physician discussions employed a mixed-methods approach. Expert review remains the gold standard, so the assessment of nine oncology and hepatology scenarios yielded 360 item-level judgments (nine dialogues × eight criteria × five raters).  The evaluation was conducted by five practicing physicians from various specialties (intensive medicine (2), general surgery, oncology, ophthalmology) and geographical locations, who also provided additional commentary.

Each dialogue was evaluated across eight criteria, organized into two main categories: Medical Content Quality and Communication Effectiveness. 
The evaluators used a 5-point Likert scale for all criteria (5 = excellent, 4 = good, 3 = acceptable, 2 = limited, and 1 = does not meet the criteria).
Although we did not use 
``LLM-as-a-Judge'' as a prompting technique, we included the evaluation criteria (see Table~\ref{tab:criteria}) in the system prompt, noted as `Reward Criteria', and asked the system to optimize the output.

\section{Results and Discussion}

\begin{figure}
  \centering
  \includegraphics[width=\linewidth]{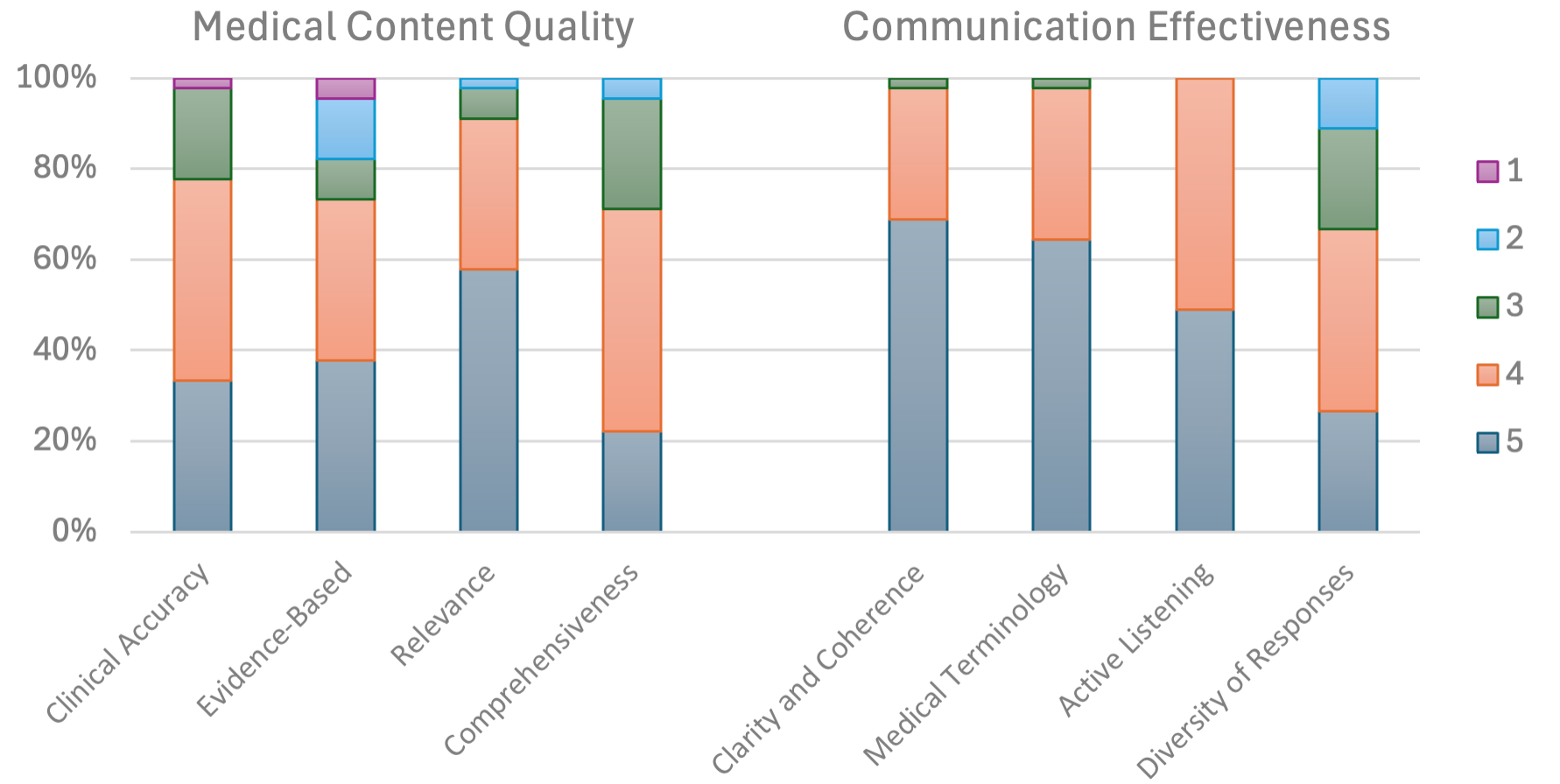}
  \caption{Evaluation by medical experts  (5=Excellent; 3=Adequate; 1=Criteria not met).}
  \label{fig:assessment}
\end{figure}

The evaluation demonstrated strong performance across both assessment categories (Fig.~\ref{fig:assessment}). Communication effectiveness achieved a mean of 4.4/5, with over 98\% of LLM-generated discussions rated ``Excellent'' (5) or ``Good'' for Clarity and Coherence, Use of Medical Terminology, and Active Listening. Medical content quality showed strong but less consistent results (mean = 4.1/5), with 91\% and 78\% of evaluations being ``Excellent'' (5) or ``Good'' (4) for Clinical Relevance and Clinical Accuracy, respectively.

To assess the agreement among physician evaluators, we employed the weighted Fleiss' $\kappa$ using quadratic weights to penalize disagreements between distant categories (e.g., 1 vs. 5), and calculated 95\% confidence intervals using bootstrap resampling with 1000 iterations.
The results show substantial agreement among the physician evaluators overall ($\kappa$ = 0.70, 95\% CI: 0.67-0.73), with also substantial agreement for both medical content quality ($\kappa$ = 0.71, 95\% CI: 0.67-0.75) and communication effectiveness ($\kappa$ = 0.68, 95\% CI: 0.64-0.72).

Our study demonstrates that large language models, when guided by a structured metadata-driven approach, can generate high-quality synthetic physician-to-physician discussions while preserving privacy. The evaluation results reveal several significant findings that advance our understanding of synthetic medical data generation.

The synthetic discussions achieved strong performance in clinical accuracy and relevance, with 91\% of evaluations rated as ``Excellent'' or ``Good'' for Clinical Relevance. This  suggests that metadata-driven generation can effectively capture the essential elements of physician-to-physician communication. However, the fact that 20\% of the discussions were rated ``Average'' or below in clinical accuracy reinforces the ongoing challenge of maintaining consistently high clinical accuracy in synthetic data generation with general purpose large language models such as GPT-4.

The framework demonstrated exceptional performance in communication effectiveness, with a mean score of 4.4/5. Particularly noteworthy was the strong performance in clarity and coherence, and appropriate use of medical terminology, with over 98\% of the evaluations rated as ``Excellent'' or ``Good'' in these areas. This suggests that the CIDI framework effectively structures synthetic medical discussions in a way that resembles real-world physician communication patterns.

Our metadata-driven approach successfully maintains privacy while generating clinically relevant discussions. By utilizing de-identified metadata rather than raw patient data as input, we address a critical challenge in AI research applied to physician-physician discussions: the need for real-world data that does not compromise patient or clinician confidentiality.

While the initial implementation of this study was conducted on proprietary LLM (GPT-4), which could potentially impact reproducibility, the CIDI framework is model-agnostic and can be implemented using various LLMs. We provide detailed prompting templates and metadata structures to enable reproduction using alternative models. Future work will explore open-source alternatives to improve accessibility and reproducibility, and/or specialized models such as MedGemma (released by Google in May 2025).

Evidence-Based Reasoning showed lower performance, with 18\% of evaluations rated below adequate level, suggesting the need for better integration of additional data sources.  This may be addressed through Retrieval-Augmented Generation (RAG). Evaluators pointed out that while some of the references provided were somewhat outdated, it is a step forward towards augmenting discussions with relevant references, which are often unavailable in live discussions. 

Moreover, although ``Variability of Responses'' had a mean of 3.8/5, this category received a rating of ``2 -Limited diversity or challenge to established ideas'' in 5 of the 45 evaluations. As some evaluators pointed out, this has been partially introduced by implementing the study with metadata mirroring exactly the real-world discussions, many of which had too few physicians engaged and/or not many responses (an evaluator also noted that not all discussions will or should lead to a wide diversity of opinions).  Therefore, this can be addressed by not limiting the number of doctors participating in the discussion and/or the number of responses when providing metadata as input.  This is an opportunity for further exploration of our framework, as increasing the variability and diversity of responses can create value for physicians and ultimately patients, as also suggested in a recent opinion paper~\cite{Sarkar2024}.

\section{Conclusion}

The proposed SynDocDis framework uses a method based on metadata from real-world patient cases to generate realistic physician-to-physician dialogues, ensuring that the generated conversations are relevant to the context, clinically accurate, and reflective of the patterns and details found in real medical discussions. By combining structured prompting techniques with privacy-preserving de-identification strategies, SynDocDis demonstrates that large language models, when guided by a structured metadata-driven approach, can ethically generate physician-to-physician discussions that are both useful and privacy-preserving.

This study offers a pathway for future research that can leverage synthetic medical dialogues for training AI assistants, refining clinical guidelines, and augmenting medical education.

\begin{credits}
\subsubsection{\ackname} 
This work was partially funded by national funds through FCT - Foundation for Science and Technology, I.P., under project UID/00127. 
We acknowledge the contributions of Dr. Einat Shacham-Shmueli, MD (Senior Oncologist and Head of the Gastrointestinal Cancer Clinic at Sheba Medical Center). We also acknowledge the input and feedback from Adir C. Sommer, MD (Department of Ophthalmology, Rambam Health Care Campus, Rambam, Israel; Ruth and Bruce Rappaport Faculty of Medicine, Technion-Israel Institute of Technology, Haifa, Israel); Alon Botzer, PhD; Amir Sheik-Yousouf, MD, MBA; Bruno Gonçalves, MD, MSc, PhD (Hospital São Lucas Copacabana);  Fabio Jung, MD, MBA; Limor Amit, MD; Ronen Tal-Botzer, PhD.

\subsubsection{\discintname}
The authors have no competing interests to declare that are relevant to the content of this article. 
\end{credits}
%
%
%
\bibliographystyle{splncs04}
\bibliography{references}

\end{document}